\documentclass{article}
\usepackage{ijcai24}

\usepackage{times}
\usepackage{soul}
\usepackage{url}
\usepackage[hidelinks]{hyperref}
\usepackage[utf8]{inputenc}
\usepackage[small]{caption}
\usepackage{graphicx}
\usepackage{amsmath}
\usepackage{amsthm}
\usepackage{booktabs}
\usepackage{algorithm}
\usepackage{algorithmic}
\usepackage[switch]{lineno}

\usepackage{multirow} 
\usepackage{amsfonts} 
\usepackage{subfig} 
\usepackage{bbding} 
\usepackage{xcolor} 

\title{Pointsoup: High-Performance and Extremely Low-Decoding-Latency Learned Geometry Codec for Large-Scale Point Cloud Scenes}

\author{
Kang You$^1$
\and
Kai Liu$^1$\and
Li Yu$^2$\and
Pan Gao$^{1}$\footnote{Corresponding author}\And
Dandan Ding$^3$
\affiliations
$^1$Nanjing University of Aeronautics and Astronautics\\
$^2$Nanjing University of Information Science and Technology\\
$^3$Hangzhou Normal University
\emails
\{youkang, liu-kai\}@nuaa.edu.cn,
li.yu@nuist.edu.cn,
pan.gao@nuaa.edu.cn,
DandanDing@hznu.edu.cn
}

\begin{document}

\maketitle

\begin{abstract}
Despite considerable progress being achieved in point cloud geometry compression, there still remains a challenge in effectively compressing large-scale scenes with sparse surfaces. Another key challenge lies in reducing decoding latency, a crucial requirement in real-world application. In this paper, we propose Pointsoup, an efficient learning-based geometry codec that attains high-performance and extremely low-decoding-latency simultaneously. Inspired by  conventional Trisoup codec, a point model-based strategy is devised to characterize local surfaces. Specifically, skin features are embedded from  local windows via an attention-based encoder, and dilated windows are introduced as cross-scale priors to infer the distribution of quantized features in parallel. During decoding, features undergo fast refinement, followed by a folding-based point generator that reconstructs point coordinates with fairly fast speed. Experiments show that Pointsoup achieves state-of-the-art performance on multiple benchmarks with significantly lower decoding complexity, i.e., up to 90$\sim$160$\times$ faster than the G-PCCv23 Trisoup decoder on a comparatively low-end platform (e.g., one RTX 2080Ti). Furthermore, it offers variable-rate control with a single neural model (2.9MB), which is attractive for industrial practitioners.
\end{abstract}


\section{Introduction}

\begin{figure}[t]
    \centering
    \includegraphics[width=0.96\linewidth]{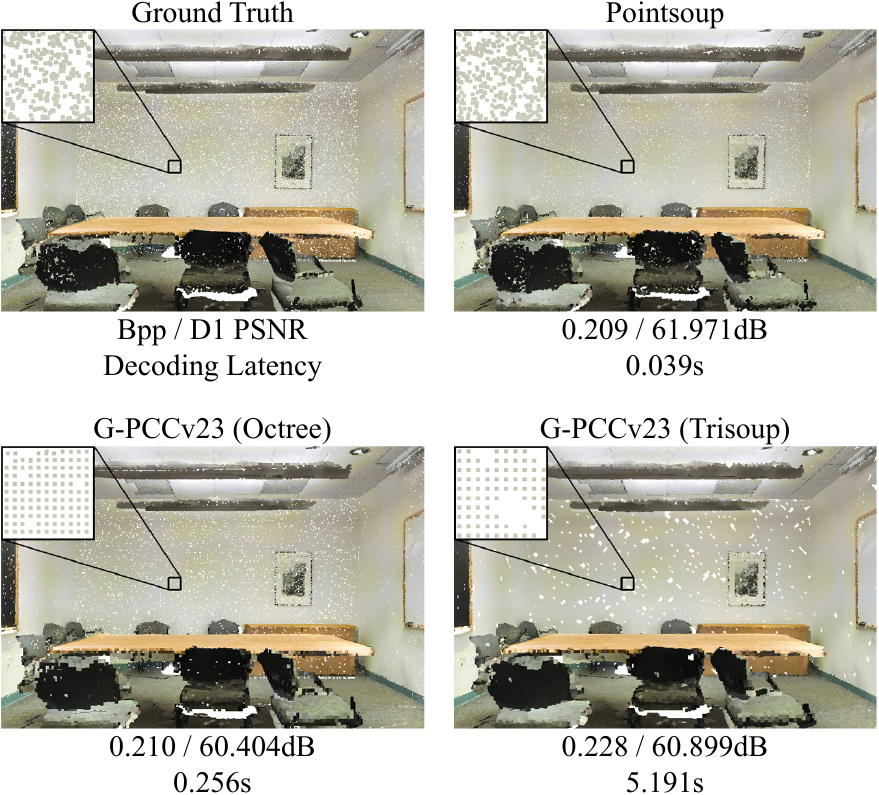}
    \caption{Quantitative compression results of proposed Pointsoup and G-PCCv23. Colors are rendered by nearest mapping. The ``conferenceRoom\_1'' in S3DIS Area 6 is used as an example, which has 1,067,709 points. Our method allows for the decoding of a million-scale point cloud geometry in 39 ms with only one \textbf{RTX 2080Ti GPU} while guaranteeing superior visual quality.}
    \label{fig:teaser}
    \vspace{-8pt}
\end{figure}

Large-scale point clouds are widely used in numerous 3D applications, such as Augmented Reality/Virtual Reality (AR/VR), autonomous driving, robotics, etc., owing to their capacity to realistically represent objects and scenes~\cite{quach2022survey,2023lidar}. A large-scale point cloud typically consists of millions of sparse points, making it challenging to store and transmit, which urges the development of Point Cloud Compression (PCC) techniques.

\textbf{Background.} Two international PCC standards, i.e., Video-based PCC (V-PCC) and Geometry-based PCC (G-PCC), were developed under the Moving Picture Experts Group (MPEG) and released as part of ISO/IEC 23090-5 and 23090-9~\cite{chen2023PCCstandards,VPCCdescription,GPCCdescription}. The octree representation~\cite{schnabel2006octree} is adopted in G-PCC to efficiently encode the geometry information. Alternatively, the Trisoup geometry codec is a viable option in G-PCC to perform lossy geometry compression more effectively by estimating local point cloud surface as triangle meshes.

Recently, owing to the significant gains obtained by learning-based approaches, both MPEG and Joint Photographic Experts Group (JPEG) committees have launched explorations on Artificial Intelligence (AI) based PCC solutions. Despite the powerful performance demonstrated in the compression of dense point cloud objects, learning-based methods~\cite{Depoco,octAttention,IPDAE,3QNet,SparsePCGC,YOGA,EHEM} still face the following two major challenges in compressing sparse point cloud scenes: 1) Unsatisfactory compression performance. The variability of sparse surfaces can make the neural model easily collapse; 2) High decoding power demands that restrict the application scope. We argue that the computational complexity for decoding should be rather low, in order to adapt to different computing-power client devices as well as live-streaming applications.

\textbf{Our Approach.} To address the above issues, we propose Pointsoup, an efficient learning-based geometry codec that attains high-performance and extremely low-decoding-latency simultaneously. Specifically, we first design the Aligned Window-based Down-Sampling (AWDS) module, which allows for the learned embedding of local surfaces, leveraging an effective attention-based aggregation. We then devise the Dilated Window-based Entropy Modeling (DWEM) module to aggregate dilated windows, which are built upon down-sampled bones, to estimate the distribution of quantized features in parallel. Finally, a fast feature refinement block is intergrated with an efficient folding-based point generator, in the Dilated Window-based Up-Sampling (DWUS) module, to reconstruct the local surface with fast speed. Moreover, the Pointsoup provides variable-rate control with a single neural network model, by fully exploiting the flexibility of the point-based pipeline.

\textbf{Contribution.} Main contributions can be summarized as:

\begin{itemize}

    \item Leveraging a point model that harnesses an effective attention-based encoder and the dilated window-based entropy modeling, our method achieves state-of-the-art compression efficiency on multiple large-scale benchmarks.
    

    \item By designing a fast feature refinement block followed by an efficient folding-based point generator, our method achieves extremely low-decoding latency. It enables nearly real-time decoding for million-scale point clouds, with up to 90$\sim$160$\times$ faster than the G-PCCv23 Trisoup decoder on only one RTX 2080Ti GPU.
    
    \item Our method shows strong generalization capability and can be readily applied to large-scale test scenarios once trained on a small-scale dataset. Furthermore, it offers flexible bitrate control with a lightweight neural model, which is beneficial for practical applications.

    
\end{itemize}


\section{Related Work}
Numerous works have contributed to the Point Cloud Geometry Compression (PCGC) task, which can be roughly divided into two categories: voxel models and point models.

\textbf{Voxel Model.} Considering the sparsity of the point cloud geometry, original Point Cloud Geometry (PCG) can be reorganized to octree structure~\cite{schnabel2006octree,octAttention,EHEM} or multi-scale sparse representation~\cite{PCGCv2,SparsePCGC,YOGA}. The octree iteratively divides the occupied space to produce an efficient tree-structured format, which is adopted by the well-known MPEG G-PCC standard~\cite{GPCCdescription} for its effectiveness and scalability. As another optional codec, Trisoup models the surface of the point cloud as a series of triangle meshes, which yields superior compression performance, but at the expense of high computational cost. The sparse tensor-based approach~\cite{SparsePCGC,YOGA,MultiscaleLiDAR} utilizes multi-scale sparse representation and delivers significant compression gains. However, stacked convolutional layers still impose substantial computational demands, limiting their application scenarios.

\textbf{Point Model.} The past several years have witnessed the emergence of point-based techniques~\cite{PCT,PTv2,vinodkumar2023survey}, which promoted the development of point models for learned PCGC tasks. Some works explored small-scale PCC techniques, but generally lack the applicability to large-scale point clouds~\cite{PDAE,PCTPCC}. Other point models for large-scale point clouds, on the other hand, struggle to achieve a competitive trade-off between compression efficiency and computing overhead~\cite{DPCC,IPDAE,3QNet,huang2023patch}. For instance, 3QNet~\cite{3QNet} devised the Model Breaking Strategy (MBS) to divide point cloud into blocks first, but the MBS can lead to significant gaps between blocks, which significantly diminishes visual quality. As a real-time codec designed for dense point maps, Depoco~\cite{Depoco} reports a low coding complexity, but it suffers from severe quality degradation.


\section{Methodology}

\begin{figure}[t]
    \centering
    \includegraphics[width=0.95\linewidth]{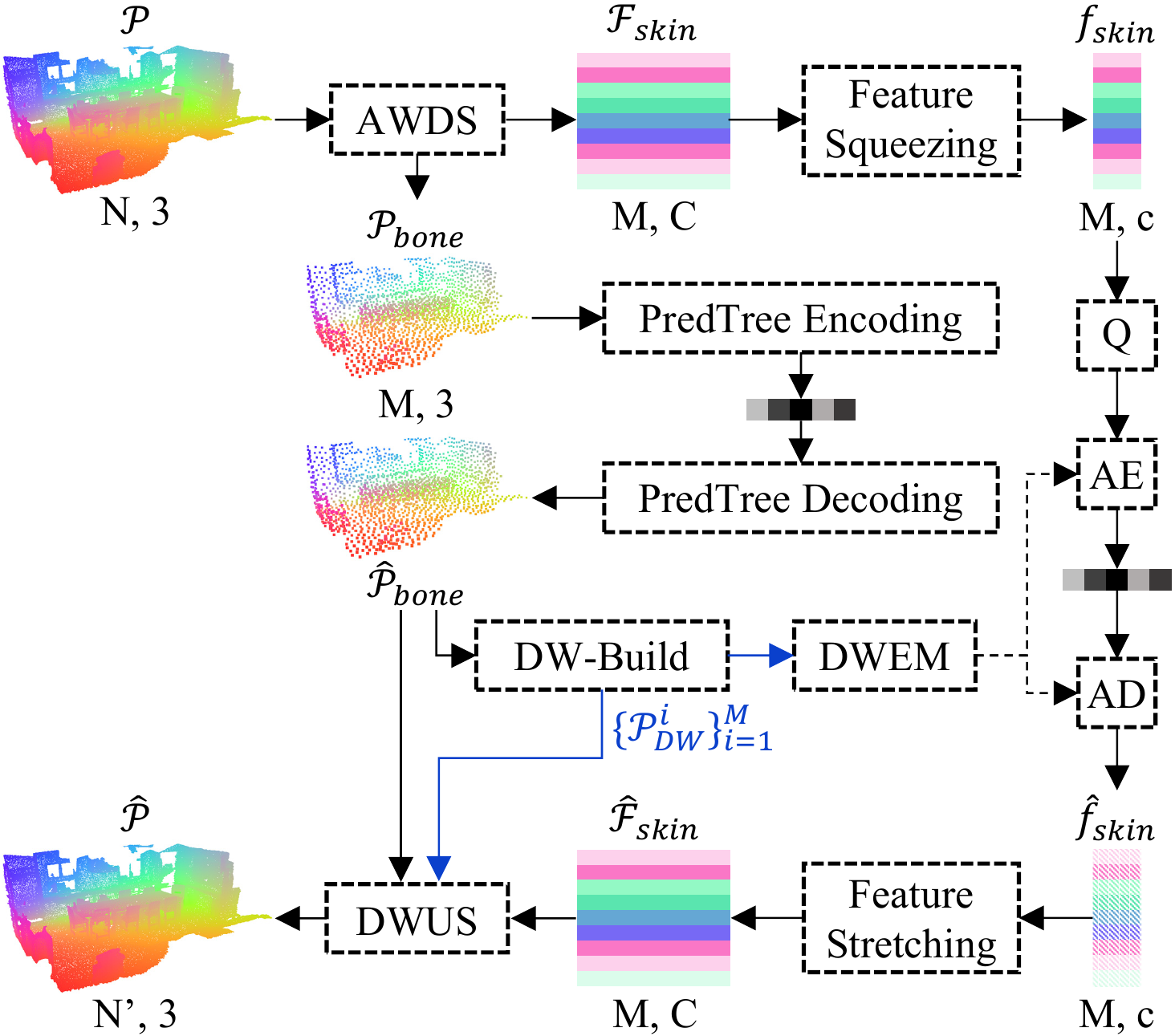}
    \caption{Pointsoup workflow. AWDS refers to the Aligned Window-based Down-Sampling module; DWEM denotes the Dilated Window-based Entropy Modeling module; DWUS represents the Dilated Window-based Up-Sampling module; AE and AD are for arithmetic encoding and decoding; Q denotes quantization.}
    \label{fig:framework}
    \vspace{-8pt}
\end{figure}

\subsection{Framework}

The overall workflow of our proposed Pointsoup is shown in Fig.~\ref{fig:framework}. Specifically, the surface of the input point cloud $\mathcal{P} \in \mathbb{R}^{N \times 3} $ is embedded to skin features $\mathcal{F}_{skin} \in \mathbb{R}^{M \times C}$ by the Aligned Window-based Down-Sampling (AWDS) module, and the down-sampled bones are then instantly encoded and decoded, through the predictive tree of G-PCC. The Dilated Window-based Entropy Modeling (DWEM) module is used to estimate the distribution of compacted skin features, by the dilated windows $\left\{\mathcal{P}^{i}_{DW}\right\}^{M}_{i=1}$ that derived from the decoded bones $\hat{\mathcal{P}}_{bone} \in \mathbb{R}^{M \times 3}$. A Dilated Window-based Up-Sampling (DWUS) module is devised to reconstruct the local surface from decoded skin features and bones. The next few subsections will detail the above-mentioned modules.

\subsection{Aligned Window-based Down-Sampling (AWDS)}
\label{sec:AWDS}

\begin{figure}[t]
    \centering
    \includegraphics[width=1.0\linewidth]{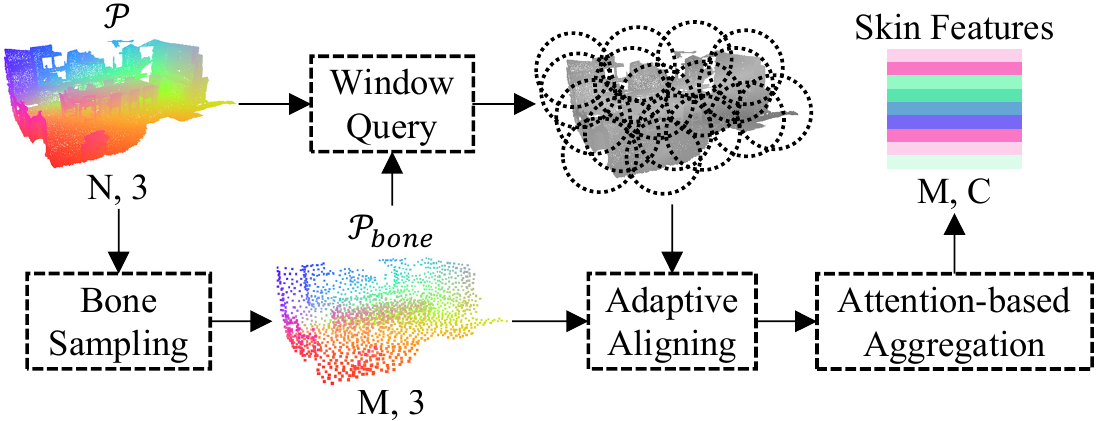}
    \caption{Aligned Window-based Down-Sampling (AWDS) module.}
    \label{fig:AWDS_flow}
    \vspace{-8pt}
\end{figure}

The AWDS module is devised to identify a well-spread skeleton and characterize the local surface into skin features, as shown in Fig.~\ref{fig:AWDS_flow}.

\subsubsection{Bone Sampling and Window Query}

The sampling and querying methods in the Pointsoup inherit the widely used aggregation basis in point cloud analysis tasks~\cite{PTv1,PointNN,ProxyFormer}, which first use the Farthest Point Sampling (FPS) to derive the skeleton $\mathcal{P}_{bone} \in \mathbb{R}^{M \times 3}$ and then build a K-Nearest Neighbor (KNN) graph to formulate overlapping local windows. Particularly, due to the high computational cost of FPS, we first use the Random Point Sampling (RPS) to obtain a subset with no more than $M\times16$ points, and then apply FPS on the subset to derive the result $\mathcal{P}_{bone}$ with $M$ points. 

\subsubsection{Adaptive Aligning}

We align the obtained overlapping local windows to facilitate network learning and enhance density adaptability. Each window is first shifted to the coordinate origin and then rescaled according to the skeleton density. Mathematically,
\begin{equation}
    d = \frac{1}{|\mathcal{P}_{bone}|} \sum_{p_{i} \in \mathcal{P}_{bone}} \min_{p_{j} \in  \mathcal{P}_{bone}} \left \{ \left \| p_{i} - p_{j} \right \| _{2} : p_{i} \ne p_{j} \right \}
\label{eq:aligning_d}
\end{equation}
\begin{equation}
    \mathcal{P}^{i}_{AW} = \left \{ \frac{p-p_{i}}{d}  : p \in \mathcal{N}(p_{i},\mathcal{P},K)) \right \}, \forall p_{i} \in \mathcal{P}_{bone}
\end{equation}
where $\mathcal{P}^{i}_{AW}$ refers to the aligned local window, and $\mathcal{N}(p_{i},\mathcal{P},K)$ represents finding $K$ nearest neighbors of the point $p_{i}$ on the input point cloud $\mathcal{P}$.

\subsubsection{Attention-based Aggregation}

\begin{figure}[t]
    \centering
    \includegraphics[width=1.0\linewidth]{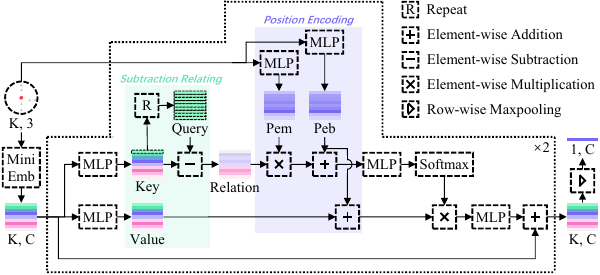}
    \caption{Attention-based aggregation of AWDS module. The self-attention block is presented in the dotted line.}
    \label{fig:AWDS_Aggregation}
    \vspace{-8pt}
\end{figure}

Given an aligned window $\mathcal{P}^{i}_{AW} \in \mathbb{R}^{K \times 3}$, we embed the local surface into a high dimensional feature vector $\mathcal{F}^{i}_{skin} \in \mathbb{R}^{1 \times C}$, by an effective attention-based neural network, as illustrated in Fig.~\ref{fig:AWDS_Aggregation}. 


Specifically, we first perform a mini embedding on each point within the window to produce feature $\mathcal{F}^{i}_{(0)} \in \mathbb{R}^{K \times C}$ for the local details, based on a pervasive graph convolution (GraphConv) operation:
\begin{equation}
    \mathcal{F}^{i}_{(0)}[j] = \text{GraphConv} \left ( \mathcal{N} \left ( p^{i}_{j}, \mathcal{P}^{i}_{AW}, k_{m} \right)  \right), \forall p^{i}_{j} \in \mathcal{P}^{i}_{AW} 
\end{equation}
where $\mathcal{N}(p_{j},\mathcal{P}^{i}_{AW},k_m)$ represents finding $k_m$ nearest neighbors of the point $p_{j}$ on the aligned window $\mathcal{P}^{i}_{AW}$; GraphConv is defined as GraphConv($\cdot$)$=$MaxPool(MLP($\cdot$)); $\mathcal{F}^{i}_{(0)}[j] \in \mathbb{R}^{1 \times C}$ refers to the $j$th feature vector (corresponding to the point $p^{i}_{j}$) in the feature matrix $\mathcal{F}^{i}_{(0)}\in \mathbb{R}^{K \times C}$.

Then, self-attention blocks are stacked following the subtraction vector attention~\cite{PTv2}. It should be noted that since the local window is generated by KNN query, the first row of the $Key$ matrix always represents the feature that attached to the center point of the window. Therefore, the $Query$ matrix is constructed by repeating the first row of the $Key$, producing the subtraction relation anchored at the window center. Then, the process of the $l$th attention block can be described as follows:
\begin{equation}
      \mathcal{S}^{i}_{(l)} = \varrho \left( \text{MLP} \left( \left( \mathcal{K}^{i}_{(l)}- \mathcal{Q}^{i}_{(l)} \right) \times Pem^{i}_{(l)} + Peb^{i}_{(l)} \right) \right)
\end{equation}
\begin{equation}
     \mathcal{F}^{i}_{(l+1)} = \mathcal{F}^{i}_{(l)} + \text{MLP} \left( \left(  \mathcal{V}^{i}_{(l)} + Peb^{i}_{(l)} \right) \times  \mathcal{S}^{i}_{(l)} \right)
\end{equation}
where $\varrho$ means the Softmax operation; $Pem$ and $Peb$ refer to the positional encoding multiplier and bias; $Query$, $Key$, and $Value$ are abbreviated as $\mathcal{Q}$, $\mathcal{K}$, and $\mathcal{V}$, for a concise explanation.

Finally, the feature $\mathcal{F}^{i}_{(L)} \in \mathbb{R}^{K \times C}$ output by the final self-attention block is aggregated to the skin feature $\mathcal{F}^{i}_{skin}\in \mathbb{R}^{1 \times C}$, by a max-pooling operation:
\begin{equation}
    \mathcal{F}^{i}_{skin} = \text{MaxPool} \left( \mathcal{F}^{i}_{(L)} \right)
\end{equation}

We set $k_m$=16 and $L$=2 in our experiment.

\subsection{Dilated Window-based Entropy Modeling (DWEM)}

\begin{figure}[t]
    \centering
    \includegraphics[width=1.0\linewidth]{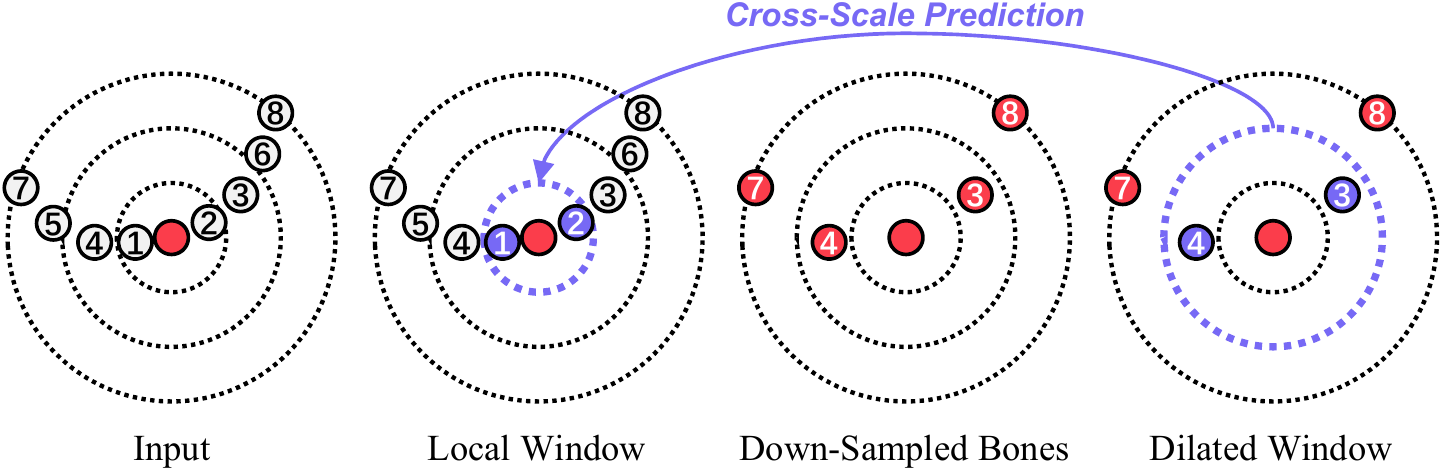}
    \caption{Dilated window-based entropy modeling. The dilated window, obtained by computing the $k$ nearest neighbors upon down-sampled bones, is introduced as a cross-scale prior.}
    \label{fig:DWEM}
    \vspace{-8pt}
\end{figure}


Considering the significant dependency among the target local window and the nearby area of the down-sampled skeleton, dilated window is introduced as the cross-scale prior to approximate the distribution of the skin feature. Meanwhile, the skin feature is further squeezed, yielding a compact representation for fast arithmetic coding.

\subsubsection{Dilated Window Construction}

As shown in Fig.~\ref{fig:DWEM}, a local window is dilated by employing the KNN graph on the down-sampled bones. Recall that the bones are handled by G-PCC at the base level, which makes dilated neighborhoods readily accessible to serve as the cross-scale prior. Mathematically, a dilated window $\mathcal{P}^{i}_{DW}$ is defined as:
\begin{equation}
    \mathcal{P}^{i}_{DW} = \mathcal{N}(p_{i},\hat{\mathcal{P}}_{bone},k), \quad \forall p_{i} \in \hat{\mathcal{P}}_{bone}
\end{equation}
where $\mathcal{P}^{i}_{DW} \in \mathbb{R}^{k \times 3}$, $\mathcal{N}$ represents the $k$ nearest neighbors of the down-sampled point $p_{i}$ on the decoded skeleton $\hat{\mathcal{P}}_{bone}$. $k$ is set to 8 in our experiments.


\subsubsection{Feature Compaction}

Higher dimensional features lead to an increase in arithmetic coding complexity, due to the presence of longer symbol sequence to be coded. Thus, a squeezing operation is used by leveraging a simple fully-connected layer. Mathematically,
\begin{equation}
    f^{i}_{skin} = \text{Linear} \left( \mathcal{F}^{i}_{skin} \right)
\end{equation}
where $\mathcal{F}^{i}_{skin} \in \mathbb{R}^{1 \times C}$, $f^{i}_{skin} \in \mathbb{R}^{1 \times c}$. Note that the skin features are stretched back by another Linear layer after arithmetic decoding. We set $C$=128 and $c$=16 in our experiments.

\subsubsection{Cross-Scale Entropy Modeling} 

A uniform scalar quantizer is used in this work, which is replaced with an additive uniform noise during training~\cite{balle2016end,jamil2023survey}. Let the quantized skin features be $\tilde{f}_{skin}=Q\left( f_{skin} \right)$, then it is further modeled as:
\begin{equation}
    P_{\theta}\left(\tilde{f}_{skin}\right) = \prod_{i=1}^{M} \left( \mathcal{L}\left( {\Phi}^{i} \right) \ast \mathcal{U}\left( -\frac{1}{2},\frac{1}{2} \right) \right) \left( \tilde{f}^{i}_{skin} \right)
\end{equation}
where $P_{\theta}$ represents the entropy model parameterized by $\theta$, $\mathcal{L}\left( {\Phi}^{i} \right)$ refers to the Laplacian distribution of quantized feature $\tilde{f}_{skin}$ with parameter ${\Phi}^{i}=({\mu}^{i},{\sigma}^{i})$, and $\mathcal{U}\left( -\frac{1}{2},\frac{1}{2} \right)$ denotes the uniform distribution ranging from $\left[-\frac{1}{2},\frac{1}{2}\right]$. Here, the parameter ${\Phi}^{i}$ can be estimated from dilated window by a network that contains a GraphConv layer and a regression head MLP:
\begin{equation}
    {\Phi}^{i} = \left({\mu}^{i},{\sigma}^{i}\right) = \text{MLP} \left( \text{GraphConv} \left( \mathcal{P}^{i}_{DW} \right) \right)
\end{equation}

Finally, the expected bit rate for the skin features can be written as:
\begin{equation}
    \mathcal{R}_{skin} = - \frac{1}{N}\log_{2}{P_{\theta}\left(\tilde{f}_{skin}\right)}
\label{eq:skin_rate}
\end{equation}
where $N$ denotes the number of points of the input point cloud.

\subsection{Dilated Window-based Up-Sampling (DWUS)}


Unlike the progressive multi-scale fashion used in the previous point models~\cite{3QNet,DPCC}, which may pose significant complexity to the decoder, we fully exploit the \emph{single-scale} strategy to reduce the computational demand: First, the skin features that are attached to the skeleton undergo fast refinement, owing to the small quantity of features $M$, which is orders of magnitude less than the original input scale $N$; Then, lightweight Folding operation is devised to generate point coordinates based on shallow MLPs.

\subsubsection{Fast Feature Refinement}

\begin{figure}[t]
    \centering
    \includegraphics[width=1.0\linewidth]{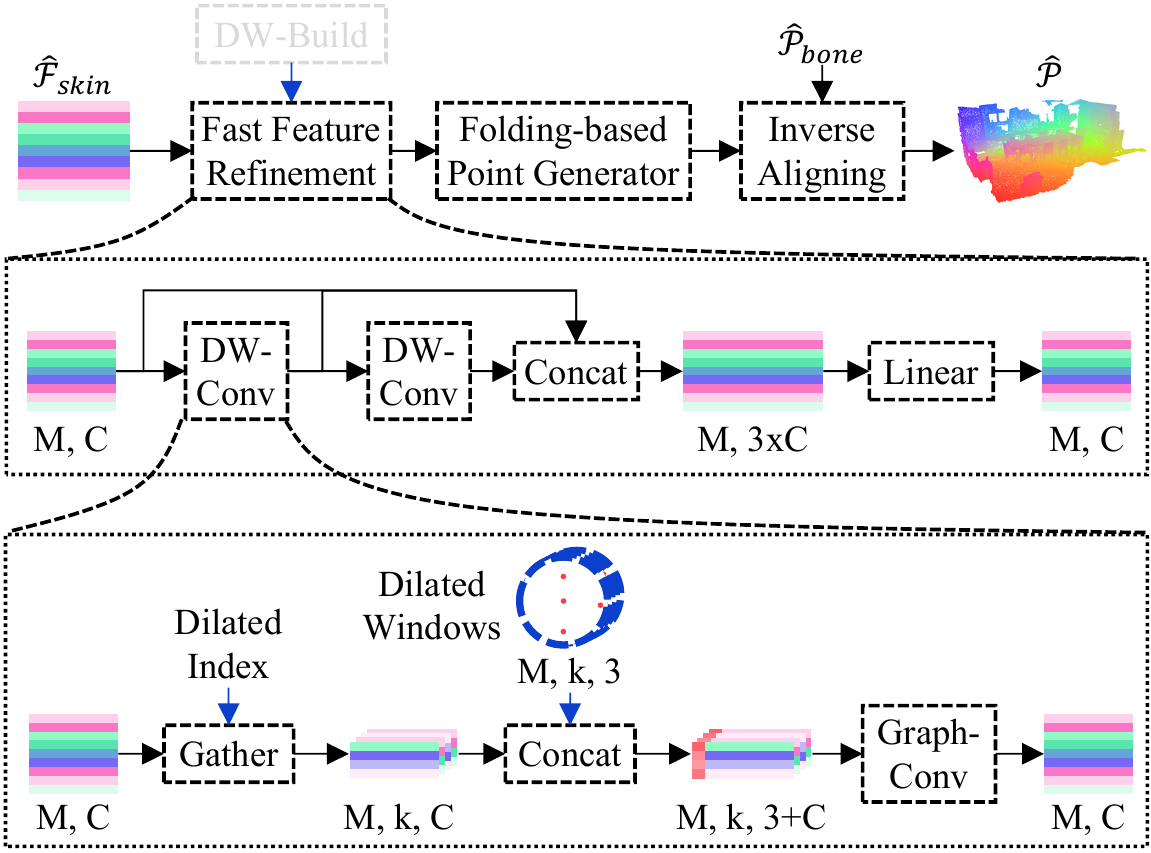}
    \caption{Dilated Window-based Up-Sampling (DWUS) module.}
    \label{fig:DWUS}
\end{figure}

Figure~\ref{fig:DWUS} details the fast feature refinement block, where the Dilated Window-based Convolution (DWConv) is introduced to integrate information from dilated windows. The reuse of the built dilated window avoids the recomputation of the spatial graph~\cite{DGCNN,li2023dynamic}, thus provides an economized graph structure for feature convolution. To be specific, features of the points within the corresponding dilated window are gathered into groups based on the provided dilated index. Then, GraphConv is applied to aggregate the gathered groups to produce refined features.

\subsubsection{Folding-based Point Generator}

\begin{figure}[t]
    \centering
    \includegraphics[width=1.0\linewidth]{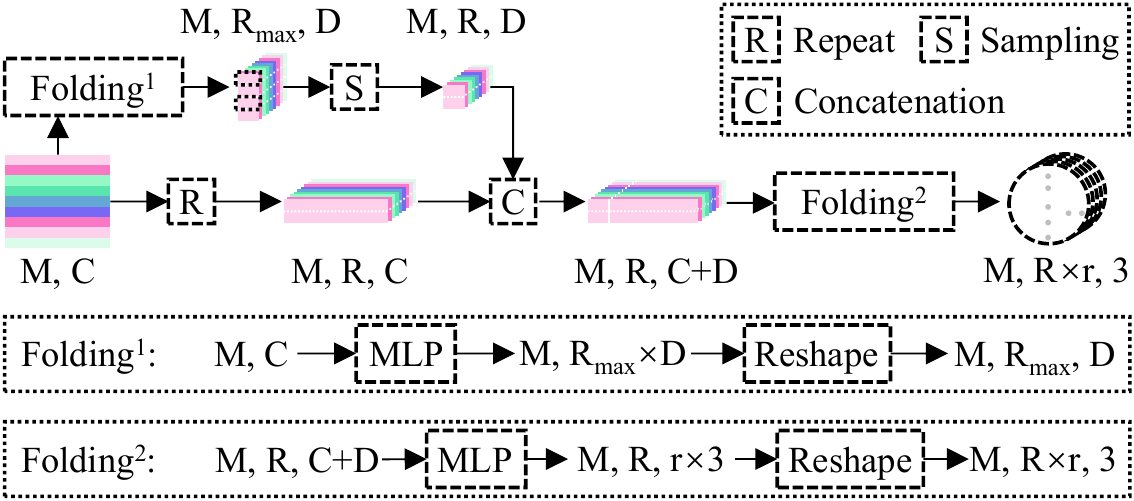}
    \caption{The proposed folding-based point generator.}
    \label{fig:point_gen}
    \vspace{-8pt}
\end{figure}

Backed by the refined features, a lightweight folding-based point generator is devised to generate point coordinates with fast speed, as shown in Fig.~\ref{fig:point_gen}. To be specific, we first define the Folding operation as a combination of MLP and reshape, where MLP is used to upscale the input features and reshape operation is used to fold the output dimensions. The first Folding transforms each skin feature from a $1 \times C$ vector to a grid matrix of $R_{max} \times D$. Then, the rows of the grid will be randomly sampled to the shape of $R \times D$, where $R \in [1, R_{max}]$. The sampling technique allows an adjustable number of generated points, which is crucial for the variable-rate control mechanism, as will be detailed in Sec.~\ref{sec:rate_control}. Then, the down-sampled feature grid is concatenated with the input skin features, followed by another Folding function to generate point coordinates.

\subsubsection{Inverse Aligning}

The inverse aligning operation mirrors the adaptive aligning used in the encoder. Each reconstructed window $\hat{\mathcal{P}}^{i}_{AW}$ is shifted to the original position and rescaled back, to assemble into a completed reconstructed result $\hat{\mathcal{P}}$. Mathematically, 
\begin{equation}
\hat{\mathcal{P}} = \bigcup_{\hat{p}_{i} \in \hat{\mathcal{P}}_{bone}}  \left \{ \left(\hat{p} \times \hat{d}\right)+\hat{p}_{i}  : \hat{p} \in \hat{\mathcal{P}}^{i}_{AW} \right \}
\end{equation}
where $\hat{d}$ represents the scale factor recalculated by $\hat{\mathcal{P}}_{bone}$, as described in Eq.~\ref{eq:aligning_d}.

\subsection{Variable-Rate Control}
\label{sec:rate_control}

A single-model-variable-rate solution is suggested based on the local window size that can be flexibly handled by the point model. Similar to the node-size adjustment of the G-PCC Trisoup codec, we modulate the size $K$ of the queried local windows to adapt to different bit rates. Inspired by~\cite{IPDAE}, we set $M= \left \lfloor \frac{2N}{K} \right \rfloor$, i.e., a denser skeleton is grown for smaller windows to capture finer details at higher bitrate budget, while a sparser skeleton is presented with larger windows to naturally reduce the bit rate. At the decoder, we employ the feature sampling technique to reconstruct the window under the given size $K$, by adapting the parameter $R$ of the point generator to $\left \lfloor \frac{K}{r} \right \rfloor$, where $r$ is fix to 4 in our implementation.


\subsection{Loss Function}

We follow the conventional rate-distortion trade-off as our loss function:
\begin{equation}
    \mathcal{L} = \mathcal{D}_{CD} ( \mathcal{P},\hat{\mathcal{P}} ) + \lambda \mathcal{R}_{skin}
\end{equation}
where $\mathcal{D}_{CD}( \mathcal{P},\hat{\mathcal{P}} )$ refers to the Chamfer Distance between input point cloud $\mathcal{P}$ and reconstructed point cloud $\hat{\mathcal{P}}$, $\mathcal{R}_{skin}$ refers to the bit rate as described in Eq.~\ref{eq:skin_rate}.

\section{Experiments}
\subsection{Experimental Setup}
\textbf{Training Dataset.} We limit the training process on the ShapeNet~\cite{chang2015shapenet} training set, which consists of 35,708 point clouds, each generated by uniformly sampling 8k points from a CAD model.

\textbf{Test Dataset.} Both large-scale indoor scenes and outdoor maps are considered for testing. The indoor point cloud scenes includes Stanford Large-Scale 3D Indoor Spaces Dataset (S3DIS)~\cite{S3DIS} and ScanNet~\cite{ScanNet}. The outdoor point cloud maps are generated from KITTI~\cite{KITTI}, following~\cite{Depoco}. Details of used test sets are provided in Tab.~\ref{tab:test_set_detail}.

{\renewcommand{\arraystretch}{0.96} 
\begin{table}[t]
    \centering
    \small
    \begin{tabular}{cccc}
    \toprule
         Dataset & Test Split & \#PC & \#Points per PC \\
    \midrule
         S3DIS & Area 6 & 48 & 3.2M$\sim$0.3M \\
         ScanNet & Official test set & 100 & 553k$\sim$32k \\
         KITTI & Sequence 08 & 186 & 554k$\sim$214k \\
    \bottomrule
    \end{tabular}
    \caption{Details for test data set used in this paper. ``PC'' denotes the abbreviation for ``Point Cloud''.}
    \label{tab:test_set_detail}
\end{table}
}

\textbf{Settings.} We implement our model using Python 3.10 and Pytorch 2.0. Adam optimizer is used with an initial learning rate of 0.0005 and a batch size of 1. We train our model only once for 140,000 steps under the local window size of 128. The $\lambda$ that balances the rate and distortion is set to $10^{-4}$. Down-sampled bones are compressed losslessly by G-PCC predictive tree. All experiments are conducted on an Intel Core i9-9900K CPU and one RTX 2080Ti GPU.

\textbf{Benchmarking Baselines.} We compare our method with state-of-the-art rules-based methods: the default Octree codec and improved Trisoup codec of the lastest G-PCCv23~\cite{GPCCdescription}, which are denoted as ``G-PCCv23'' and ``G-PCCv23 (T)'', respectively; and learning-based methods: OctAttention~\cite{octAttention}, IPDAE~\cite{IPDAE}, and 3QNet~\cite{3QNet}. All learning-based methods are retrained on the same dataset as our method, and all test samples are normalized to the coordinate range of [0, 1023] (a.k.a., 10-bit precision) for ease of fair comparison. In addition, we compare Depoco~\cite{Depoco} and D-PCC~\cite{DPCC} following their recommended test conditions in Sec.~\ref{sec:scenario_dependent}.

\subsection{Quantitative Comparison}
\label{sec:comparison}

{\renewcommand{\arraystretch}{0.96} 
\begin{table*}[t]
    \centering
    \small
    \begin{tabular}{ ccccccccc }
    \toprule
         Dataset & Metric & G-PCCv23 & G-PCCv23 (T) & OctAttention & IPDAE & 3QNet & Pointsoup \\
    \midrule
        \multirow{2}{*}{S3DIS} & BD-PSNR (dB) & - & \underline{+2.256} & -1.085 & -0.745 & -2.461 & \textbf{+3.229} \\
        & BD-Rate (\%) & - & \textbf{-67.712} & +38.385 & +25.838 & +138.569& \underline{-60.067} \\
    \midrule
        \multirow{2}{*}{ScanNet} & BD-PSNR (dB) & - & \underline{+2.477} & +0.234 & +0.613 & -0.233 & \textbf{+4.195} \\
        & BD-Rate (\%)) & - & \underline{-46.310} & -6.038 & -14.023 & +4.122 & \textbf{-60.302} \\
    \midrule
        \multirow{2}{*}{KITTI} & BD-PSNR (dB) & - & \underline{+3.093} & -0.430 & -2.222 & -0.891 & \textbf{+3.392} \\
        & BD-Rate (\%) & - & \underline{-61.517} & +12.344 & +87.928 & +27.486 & \textbf{-64.105} \\
    \midrule
        \multicolumn{2}{c}{Avg. Time (s/frame)} & Enc / Dec & Enc / Dec & Enc / Dec & Enc / Dec & Enc / Dec & Enc / Dec \\
    \midrule
        \multicolumn{2}{c}{S3DIS} & \textbf{0.334} / \underline{0.126} & 11.477 / 2.143 & \underline{0.672} / 92.055 & 20.846 / 1.224 & 24.248 / 0.401 & 8.149 / \textbf{0.022} \\
    \midrule
        \multicolumn{2}{c}{ScanNet}  & \textbf{0.060} / \underline{0.027} & 7.365 / 1.122 & \underline{0.109} / 12.807 & 4.641 / 0.237 & 2.255 / 0.156 & 2.833 / \textbf{0.006} \\
    \midrule
        \multicolumn{2}{c}{KITTI} & \textbf{0.110} / \underline{0.048} & 8.508 / 1.819 & \underline{0.231} / 26.658 & 6.944 / 0.459 & 5.865 / 0.258 & 3.236 / \textbf{0.011} \\
    \bottomrule
    
    \end{tabular}
    \caption{Quantitative results using BD-PSNR and BD-Rate metrics. G-PCCv23 serves as the anchor. The \textbf{best} and \underline{second-best} results are highlighted in bold and underlined, respectively.}
    \label{tab:scene_results}
    \vspace{-4pt}
\end{table*}
}

\begin{figure}[t]
    \centering
    \includegraphics[width=1.0\linewidth]{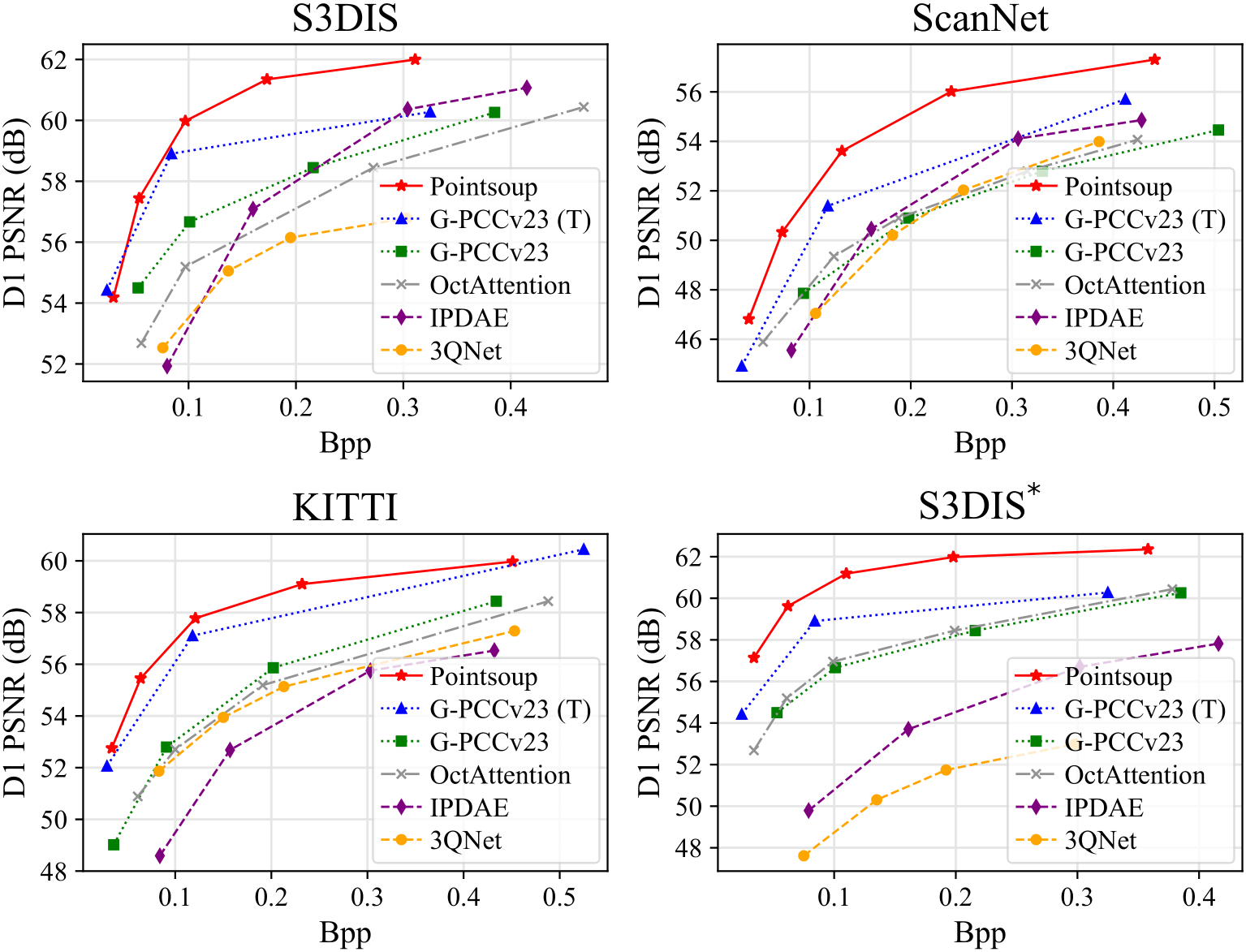}
    \caption{Rate-distortion performance comparison on S3DIS, ScanNet, and KITTI. ``S3DIS$^{\ast}$'' refers to the evaluation of the models that are trained on Area 1$\sim$5, instead of ShapeNet, and tested on Area 6.}
    \label{fig:scene_results}
    \vspace{-8pt}
\end{figure}

\begin{figure}[t]
    \centering
    \includegraphics[width=1.0\linewidth]{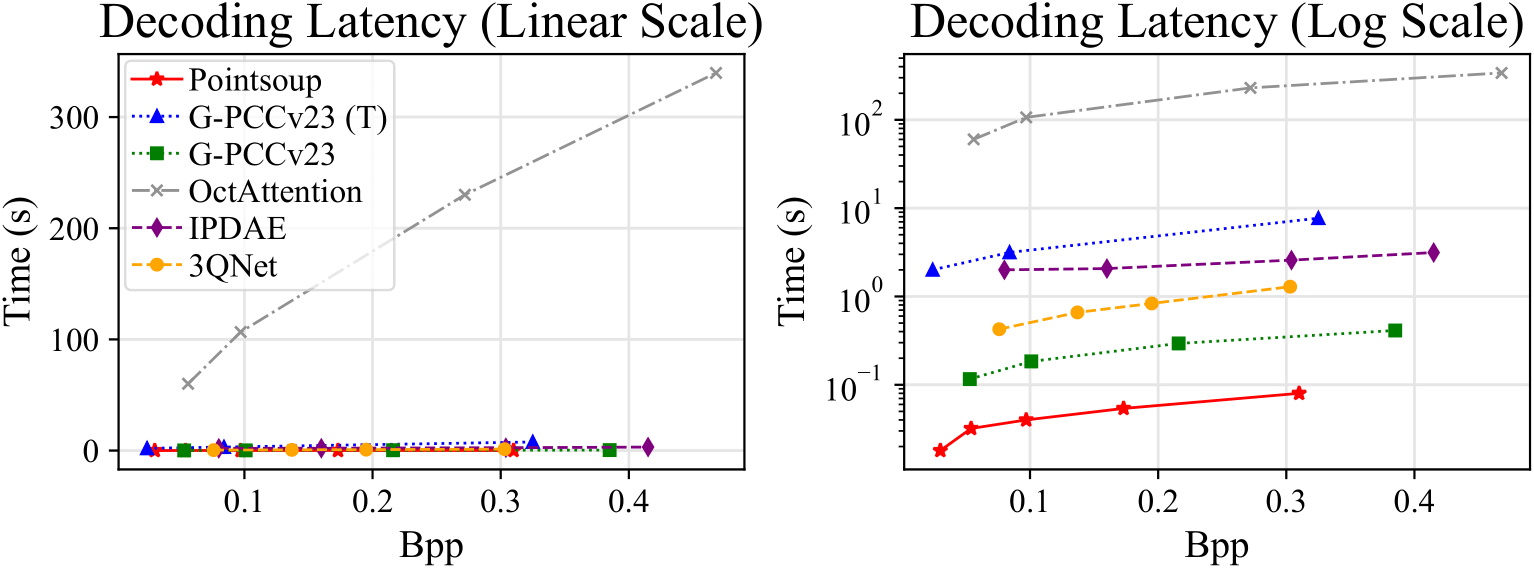}
    \caption{Decoding time comparison. We illustrate the time consumption at different bit rates, where each data point refers to the average decoding time for all test scenes in S3DIS Area 6.}
    \label{fig:decoding_latency}
    \vspace{-4pt}
\end{figure}

\begin{figure}[t]
    \centering
    \includegraphics[width=1.0\linewidth]{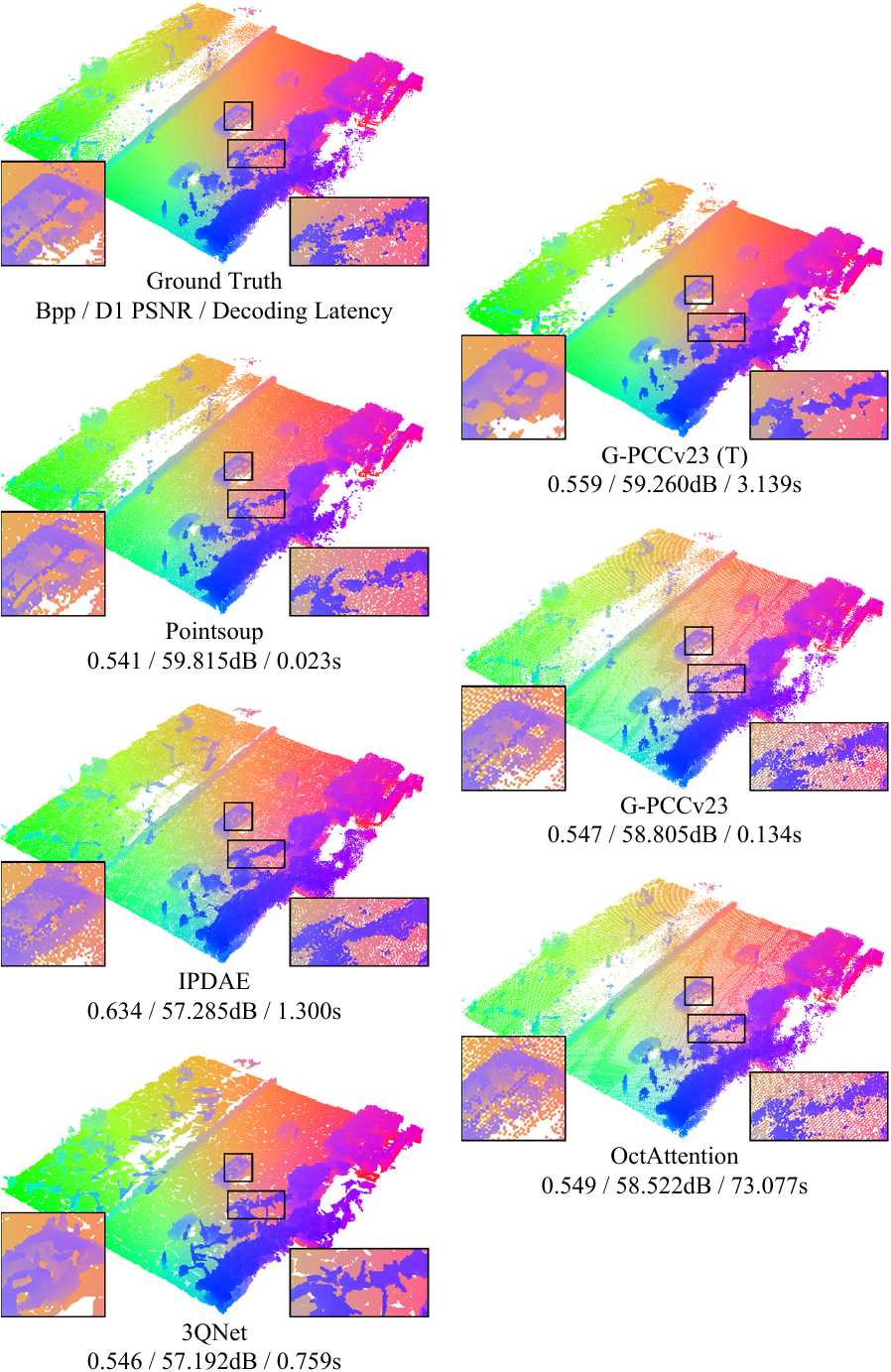}
    \caption{Reconstruction visualization of an example dense point cloud map in KITTI sequence 08.}
    \label{fig:visualized_dense_map}
\end{figure}

\textbf{Rate-Distortion Performance.}
Figure~\ref{fig:scene_results} shows the rate-distortion curves of different methods and Tab.~\ref{tab:scene_results} demonstrates the quantitative results using BD-PSNR and BD-Rate metrics. It can be seen that the proposed Pointsoup achieves the best rate-distortion performance, providing 60\%$\sim$64\% bitrate reduction over the G-PCCv23 anchor.

\textbf{Subjective Visual Quality.} Figure~\ref{fig:teaser} shows visualized results for an example indoor scene, where our method reconstructs the point cloud surface with a uniform point distribution that shares the same nature as the ground truth. An example of an outdoor point cloud map is presented in Fig.~\ref{fig:visualized_dense_map}, in which finer details can be observed in our output (e.g., higher fidelity vehicle shapes and vegetation outlines).

\textbf{Computational Complexity.} As can be seen from Fig.~\ref{fig:decoding_latency} and Tab.~\ref{tab:scene_results}, our method reports the lowest decoding latency, e.g., 90$\sim$160$\times$ faster than the G-PCCv23 Trisoup decoder and 3$\sim$5$\times$ faster than the Octree decoder. Moreover, the encoding time of the proposed Pointsoup is also significantly faster than the Trisoup, presenting a manageable encoding complexity. In addition, our network is fairly small with 761k parameters (about 2.9MB), which is much lighter than other point models such as 3QNet (85MB) and IPDAE (68$\sim$516MB for each bitrate point).

\subsection{Customized Training Domain}

The training process in the above section is limited to the ShapeNet dataset. Intuitively, it is worth considering a training domain that closely resembles the test scenario to enhance reconstruction accuracy. Therefore, we retrain each model on S3DIS Area 1$\sim$5 and test them on Area 6 again. As seen from Fig.~\ref{fig:scene_results} (S3DIS$^{\ast}$), the performances of Pointsoup and OctAttention are significantly improved, due to the similar patterns shared between training and test samples. Unexpectedly, both IPDAE and 3QNet exhibit a severe degradation, possibly indicating their inadequate capacity in handling complex training samples~\cite{CLsurvey}.

\subsection{Scenario Dependent Comparison}
\label{sec:scenario_dependent}

Depoco~\cite{Depoco} and D-PCC~\cite{DPCC} are representative point models for large-scale point cloud compression. However, they do not support the generalization from a small-scale training set (e.g., ShapeNet) to large-scale test frames. In this subsection, we follow their training/testing conditions for a comparative study, i.e., we verify our model using the respective training and test datasets as suggested in their papers. As seen from Tab.~\ref{tab:depoco_comparison}, our method significantly outperforms Depoco and D-PCC in terms of rate-distortion performance. It is worth noting that, despite the faster coding speed of Depoco, it comes at the expense of severely compromised reconstruction quality.

{\renewcommand{\arraystretch}{0.96} 
\setlength{\tabcolsep}{0.46em} 
\begin{table}[t]
    \small
    \centering
    \begin{tabular}{lrrrr}
    \toprule
         \multirow{2}{*}{Metric} & \multicolumn{2}{c}{Conditions of Depoco} & \multicolumn{2}{c}{Conditions of D-PCC} \\
         \cmidrule{2-5}
         & Pointsoup & Depoco & Pointsoup & D-PCC \\
    \midrule
        BD-PSNR (dB)     & \textbf{+3.392} & -2.061 & \textbf{+4.180} & -5.480 \\
        BD-Rate (\%)     & \textbf{-64.105} & +54.037 & \textbf{-65.682} & +165.138  \\
    \midrule
    Enc. Time (s) & 3.234 & \textbf{0.131} & 1.480 & \textbf{0.646} \\ 
    Dec. Time (s) & 0.011 & \textbf{0.002} & \textbf{0.006} & 0.165 \\
    \bottomrule
    \end{tabular}
    \caption{Compression efficiency comparison on the test conditions of Depoco and D-PCC. G-PCCv23 serves as the anchor.}
    \label{tab:depoco_comparison}
\end{table}
}

Please refer to our supplementary material for more comparisons and qualitative results.

\subsection{Ablation Study}

\begin{figure}[t]
    \centering
    \includegraphics[width=1.0\linewidth]{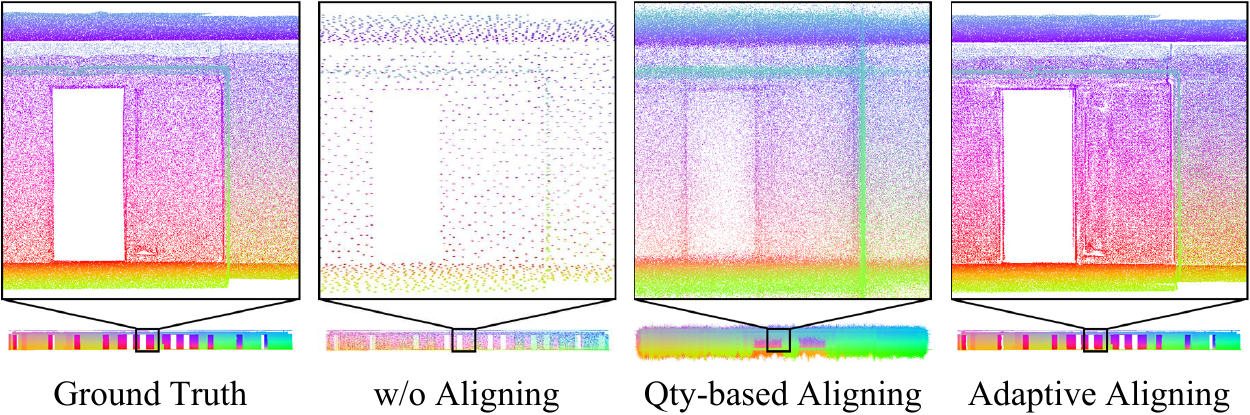}
    \caption{Comparison of proposed adaptive aligning and quantity (Qty)-based aligning.}
    \label{fig:ablation_aligning}
\end{figure}

\begin{figure}[t]
    \centering
    \includegraphics[width=1.0\linewidth]{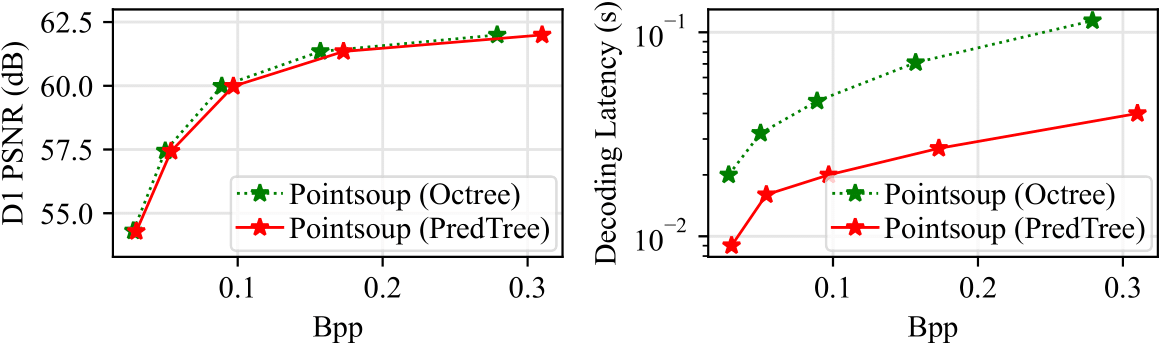}
    \caption{Comparison between using predictive tree and octree codec for bones. S3DIS Area 6 is used for test.}
    \label{fig:ablation_predtree}
    \vspace{-6pt}
\end{figure}

{\renewcommand{\arraystretch}{0.96} 
\begin{table}[h!]
    \small
    \centering
    \begin{tabular}{ cccrr }
    \toprule
        ATTN & DWEM & FFR & BD-PSNR (dB) & BD-Rate (\%)  \\
    \midrule
        \XSolidBrush & \CheckmarkBold & \CheckmarkBold & -0.385 & +14.647 \\
        \CheckmarkBold & \XSolidBrush & \CheckmarkBold & -1.128 & +47.243 \\
        \CheckmarkBold & \CheckmarkBold & \XSolidBrush & -0.210 & +7.588 \\
    \bottomrule
    \end{tabular}
    \caption{Ablation study for network components. ``ATTN'' refers to the attention block used in the aggregation of the AWDS module. ``DWEM'' refers to the dilated window-based entropy modeling. ``FFR'' refers to the fast feature refinement block of the DWUS module. Fully armed Pointsoup serves as the anchor. Models are tested on S3DIS.}
    \label{tab:ablation_network}
    \vspace{-6pt}
\end{table}
}

{\renewcommand{\arraystretch}{0.96} 
\setlength{\tabcolsep}{0.46em} 
\begin{table}[b!]
    \small
    \centering
    \begin{tabular}{lrrrr}
    \toprule
        & BD-Rate (\%) & Dec (ms) & AD (ms) \\
    \midrule
        w/o Feature Compaction & -54.820 & 42 & 22\\
        Pointsoup  & \textbf{-60.067} & \textbf{22} & \textbf{4} \\
    \bottomrule
    \end{tabular}
    \caption{Ablation study on feature compaction. S3DIS Area 6 is used for test. G-PCCv23 serves as the anchor. AD refers to the time of arithmetic decoding.}
    \label{tab:ablation_compaction}
\end{table}
}

\textbf{Adaptive Aligning.} It is imperative to consider the adaptation of the point densities as they vary with the number of points and the size of the scenes. The quantity-based aligning~\cite{IPDAE} is another reasonable way of density adaptation. However, they only consider the influence of the number of points, while neglecting the impact of the spatial volume of the point cloud. For instance, they do not adapt well to narrow hallways, as evidenced in Fig.~\ref{fig:ablation_aligning}. On the contrary, we infer local densities from down-sampled bones, which provides better aligned results by exploiting cross-scale priors.

\textbf{Predictive Tree vs. Octree.} Octree is another optional codec to compress down-sampled bone points. However, since predictive tree is particularly designed for low complexity decoding, using the octree will decelerate the decoding speed (about 2$\sim$3$\times$ slower), as shown in Fig.~\ref{fig:ablation_predtree}, despite some rate-distortion benefits (7.979\% BD-BR gain).


\textbf{Network Components.} As shown in Tab.~\ref{tab:ablation_network}, several key modules are disabled individually to examine the validity of the components. Note that the attention block is changed to MLP, and the DWEM is replaced by the basic factorized prior model~\cite{balle2016end} during study. Results show that disabling the DWEM module will lead to 47.243\% BD-Rate loss relative to the original Pointsoup, which demonstrates the significant efficiency of the cross-scale entropy modeling.

\textbf{Feature Compaction.} We use a linear layer to squeeze skin feature to a compact representation, which speed up the arithmetic coding of the features. As shown in Tab.~\ref{tab:ablation_compaction}, the feature compaction operation greatly improves the speed of the decoding, reducing the arithmetic decoding time from 22 ms to 4 ms. In addition, the compaction yields a more efficient representation, leads to a slight bitrate reduction.

Please refer to our supplementary material for more ablation studies and discussions.

\section{Conclusion}

This paper proposes an efficient learning-based geometry codec, dubbed Pointsoup, aiming at large-scale point cloud scenes. The Pointsoup demonstrates state-of-the-art compression efficiency with significantly low decoding latency and variable-rate controllability, making it a promising option for AI-based PCC solutions. Downstream tasks (e.g., object detection) may consider working directly on compressed domain without the need of a complete reconstruction. Source code and supplementary material are available at \url{https://github.com/I2-Multimedia-Lab/Pointsoup}.

\appendix



\section*{Acknowledgments}

This work was supported by the National Natural Science Foundation of China (No. 62272227).

\bibliographystyle{named}
\bibliography{ijcai24}

\end{document}